\documentclass[runningheads]{llncs}

\usepackage{easyReview}

\usepackage{cite}
\usepackage{url}
\usepackage{babel}
\usepackage{amsmath}
\usepackage{graphicx}

\usepackage{longtable}
\usepackage{floatrow}
\usepackage{subfig}
\usepackage{multirow}
\usepackage{dcolumn}
\usepackage[font=small,labelfont=bf]{caption}
\usepackage{algpseudocode} 
\usepackage[ruled,linesnumbered,lined,boxed,commentsnumbered]{algorithm2e}
\usepackage{algpseudocode}
\usepackage{tabularx}
\usepackage{adjustbox}
\usepackage{mathabx}
\usepackage{array}
\usepackage{mdwmath}
\usepackage{mdwtab}
\usepackage{eqparbox}

\usepackage{balance}  

\usepackage{caption}
\begin{document}

\title{Towards Synthetic Multivariate Time Series Generation for Flare Forecasting}

\author{Yang Chen\thanks{Corresponding author: Yang Chen \\ ychen113@student.gsu.edu} \and
        Dustin J. Kempton \and Azim Ahmadzadeh \and Rafal A. Angryk
    }
\authorrunning{Yang Chen et al.}
%
\institute{
        Georgia State University \\
        Atlanta, GA 30302, USA
    } 

\maketitle

\begin{center}
        (Accepted May 15, 2021) \\
        \medskip
        Submitted to ICAISC 2021
\end{center}


\begin{abstract}

    One of the limiting factors in training data-driven, rare-event prediction algorithms is the scarcity of the events of interest resulting in an extreme imbalance in the data. There have been many methods introduced in the literature for overcoming this issue; simple data manipulation through undersampling and oversampling, utilizing cost-sensitive learning algorithms, or by generating synthetic data points following the distribution of the existing data. While synthetic data generation has recently received a great deal of attention, there are real challenges involved in doing so for high-dimensional data such as multivariate time series. In this study, we explore the usefulness of the conditional generative adversarial network (CGAN) as a means to perform data-informed oversampling in order to balance a large dataset of multivariate time series. We utilize a flare forecasting benchmark dataset, named SWAN-SF, and design two verification methods to both quantitatively and qualitatively evaluate the similarity between the generated minority and the ground-truth samples. We further assess the quality of the generated samples by training a classical, supervised machine learning algorithm on synthetic data, and testing the trained model on the unseen, real data. The results show that the classifier trained on the data augmented with the synthetic multivariate time series achieves a significant improvement compared with the case where no augmentation is used. The popular flare forecasting evaluation metrics, TSS and HSS, report 20-fold and 5-fold improvements, respectively, indicating the remarkable statistical similarities, and the usefulness of CGAN-based data generation for complicated tasks such as flare forecasting.
    
    \begin{keywords}
    multivariate time series, class imbalance, generative adversarial network, flare forecasting
    \end{keywords}
    
\end{abstract}


\section{Introduction}

    In February 2010, NASA launched the Solar Dynamics Observatory, the first mission of NASA's Living with a Star program, which is a long term project dedicated to the study of the Sun and its impacts on human life \cite{withbroe:lws}. The SDO mission is an invaluable instrument for researching solar activity, which can produce damaging space weather. This space weather activity can have drastic impacts on space and air travel, power grids, GPS, and communications satellites \cite{nrc:spaceweather}. For example, in March of 1989, geomagnetically induced currents, produced when charged particles from a coronal mass ejection impacted the earth's atmosphere, caused power blackouts and direct costs of tens of millions of dollars to the electric utility ``Hydro-Qubec'' \cite{boteler:geomagnethazards}.  If a similar event would have happened during the summer months,  it is estimated that it would likely have produced widespread blackouts in the northeastern United States, causing an economic impact in the billions of dollars \cite{boteler:geomagnethazards}.
    
    A solar flare is an event occurring in the solar corona that is characterized by a sudden orders-of-magnitude brightening in Extreme Ultra-Violet (EUV) and X-ray, and for large events, gamma-ray emissions, from a small area on the Sun, lasting from minutes to a few hours \cite{Benz2008, martens2017data}. The classification system for solar flares are on a logarithmic scale and uses the letters A, B, C, M or X, according to the peak X-ray flux. In a typical binary classification strategy, M and X classes are identified as the positive class while no flare occurrence and flares of A, B and C classes are identified as the negative class.
    
    The goal of this project is to generate synthetic data, especially for multivariate time series of magnetic filed parameters leading up to solar flares. As discussed in \cite{max2019understanding,ahmadzdeh2021how}, the extreme class-imbalance between positive and negative classes in the solar flare data, and the improper treatment of said extreme imbalance between classes, can result in unrealistic and unreliable analyses, with little practical value in flare forecasting, flare classification, and flare clustering. Therefore, solving the issue of insufficient positive class of flare data is an important problem for current research in this domain. As such, this project is dedicating to generate realistic flare data based on real data, in order to provide a balanced training dataset for use in such problems.


\section{Related work}
    
    Although remedies such as oversampling, undersampling, and cost-sensitive learning, have been considered to tackle this problem \cite{ahmadzadeh:challenges,ahmadzadeh:rareevent}, these methods can only provide limited improvements since they do not introduce or utilize any new data. The development of generative modeling provides an attractive alternative and potentially more domain-specific approach for data augmentation. For example, the Generative Adversarial Network can be trained to learn data distributions of the minority classes, thereby generating synthetic data for constructing a balanced and larger dataset to train more unbiased and powerful classifiers.

    \subsection {Generative Adversarial Network (GAN)}\label{sect:related-work:gan}

    First proposed in \cite{goodfellow2014generative}, the Generative Adversarial Network (GAN) tries to learn an implicit density of real samples. The GAN trains the two components in an adversarial way. First, the generator is used to sample initial inputs from a latent space, which is used to produce data similar to real data. Next step, both generated samples and real data are used as inputs of a discriminator, and the discriminator assigns the label of samples after processing inputs with a neural network. Eventually, the predicted labels are used to calculate errors with a defined objective function, and the result is used for adjusting the whole model. This mechanism can help a generator gradually generate better realistic samples under the supervision of real samples, and this process keeps running until the discriminator cannot distinguish real data and synthetic samples. 
    
    There have been various types of GANs proposed as an extension of the vanilla GAN to deal with different demands. For instance, in the computer vision domain, the Deep Convolutional GAN \cite{Radford2016UnsupervisedRL} has been applied to learn reusable feature representations and generate synthetic images by utilizing convolutional neural networks as the generator and discriminator. The Wasserstein GAN \cite{arjovsky2017wasserstein} commits to improve the stability of learning and provide a meaningful learning curve. The Info GAN \cite{chen2016infogan} incorporates the representation learning by encoding features into the latent vector. The Conditional GAN (CGAN) \cite{mirza2014conditional} is dedicated to improving the quality of generated samples and controlling the classes of synthetic samples by utilizing conditional information. Finally, the advantage of controlling the mode of generated samples makes CGAN become the most appropriate framework in our study since our goal is to generate synthetic samples of multiple minority classes for tackling the class imbalance issue.
    
    \subsection{Time series generation}

    There are already several projects in different domains that have worked on generating time series data utilizing the Generative Adversarial Network. In \cite{esteban2017real}, the RGAN was utilized to generate medical time series data implemented with Long Short-Term Memory (LSTM) network. The motivation of their work was to develop a privacy-preserving method of generating synthetic medical data for machine learning modeling since actual patient data is sensitive to privacy issues. In \cite{mogren2016c}, the use of a C-RNN-GAN was proposed as a method to generate musical data. This method differs from \cite{esteban2017real}, in that it applied a unidirectional LSTM for the generator and a bidirectional LSTM for the discriminator. Then, in \cite{yoon2019time}, the TimeGAN was proposed that combines the versatility of the unsupervised GAN approach with the control over conditional temporal dynamics. This method has two more autoencoding components, including an embedding function and a recovery function trained jointly with generator and discriminator components. This structure enables the model can learn to encode features, generate representations and iterate across time simultaneously.

    \subsection{Multivariate time series dataset}\label{sect:related-work:swan-sf}

    The data primarily used in this project is a benchmark dataset, named as Space Weather ANalytics for Solar Flares (SWAN-SF), recently released by \cite{angryk2020multivariate}. SWAN-SF is a comprehensive, multivariate time series (MVTS) dataset extracted from solar photospheric vector magnetograms in HMI Active Region Patch (HARP) data made available as the Spaceweather HMI Active Region Patch (SHARP) series \cite{hoeksema2014, bobra2014helioseismic}. The SWAN-SF is made up of five temporally non-overlapping partitions covering the period from May 2010 through August 2018. Each partition contains approximately an equal number of X- and M-class flares, and there is a total of 6,234 flare records and 324,952 no-flare records. Comparing the amount of two kinds of data, we can find an extremely imbalanced issue in this dataset. As mentioned previously, \cite{ahmadzdeh2021how} showed that the extreme class imbalance between positive and negative classes in the solar flare data, and the improper treatment of said extreme imbalance can result in unrealistic and unreliable analyses. Furthermore, each flare record is a multivariate time series with 60 time steps, and each time step has 51 magnetic field parameters. This work will focus on four parameters, including TOTUSJH, ABSNJZH, SAVNCPP, and TOTBSQ, as a representative subset of the full 51 field parameters (for the definition of parameters see Table 1 in \cite{angryk2020multivariate}). This is because many of the parameters are highly correlated, leading most studying flare forecasting to utilize some subset of the full set.

\section{Methodology} \label{sect:methodology}

\begin{figure} [t!]
      \centering
    \includegraphics[width=\textwidth]{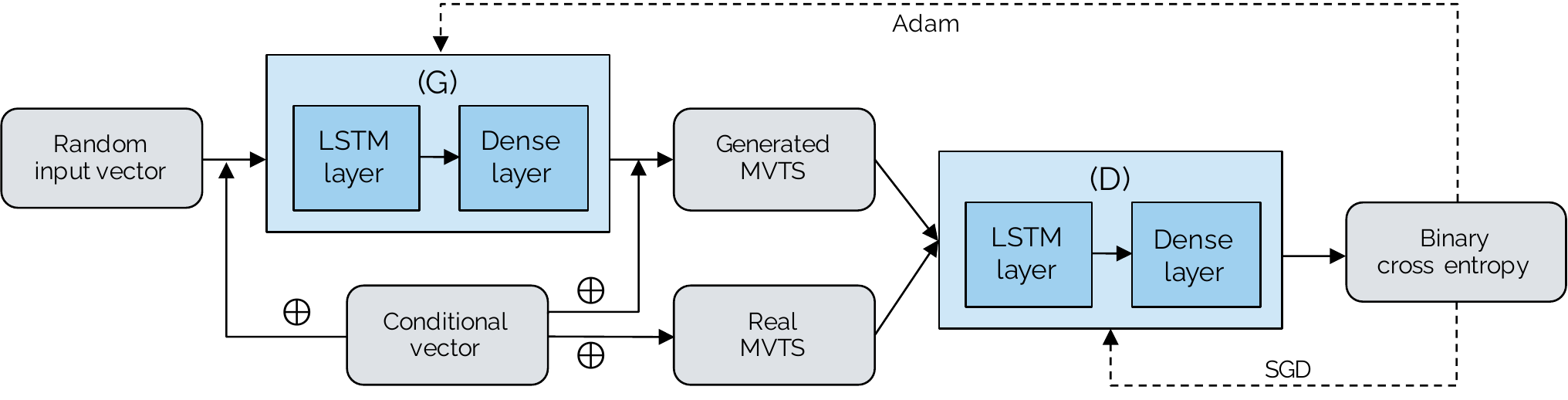}
      \caption{This is the framework of the CGAN model, including components of the generator (\textit{G}) and the discriminator (\textit{D}). Each component is processed by the combination of the LSTM layer and the Dense layer. The inputs of the generator are random input vectors concatenated with conditional vectors. The inputs of the discriminator are either synthetic or real multivariate time series with conditional vectors. The binary cross-entropy is the criterion for optimizing the model.}
      \label{fig:cgan_framework}
    \end{figure}

    We decide to use the Conditional Generative Adversarial Network (CGAN) in this study for several reasons. First, the advantage of controlling the mode of generated samples allows us to generate samples of minority classes to tackle the class imbalance issue. Second, CGAN can provide a stable and faster training compared to the vanilla GAN. Moreover, the SWAN-SF dataset is labelled, therefore it can provide conditional information for us to train CGAN models. LSTM network is utilized as the basic components in both the generator and the discriminator illustrated in Fig.~\ref{fig:cgan_framework} since we are processing time series data.

    The ultimate goal of a generator \textit{G} is to generate an output with similar characteristics as the real data. As seen in Fig.~\ref{fig:cgan_framework}, the model takes in a random input vector $Z_{n}$, which is a tensor with the shape of [${batch\_size}$, ${sequence\_length}$, ${latent\_dim}$]. In this study, the shape is $[32, 60, 3]$ for $32$ multivariate time series in a batch, each of length $60$ and latent dimensions $3$. Moreover, the conditional vector, namely $C_{n}$, has the shape of $[32, 60, 2]$ since it is encoded in one-hot representation with labels of binary classes. Finally, we concatenate $Z_{n}$ and $C_{n}$, obtaining a tensor of shape $[32, 60, 5]$ as the final input of the generator. After going through the calculations of LSTM layer and Dense layer, the outputs, regarded as synthetic samples, have the same shape as the real data, i.e., $[32, 60, 4]$ where $4$ stands for four magnetic field parameters mentioned in \ref{sect:related-work:swan-sf}.
    
    The task of a discriminator \textit{D} is to classify inputs as either being real or synthetic samples generated by the generator. In Fig.~\ref{fig:cgan_framework}, it can be seen that the discriminator takes two forms of multivariate time series (MVTS) as the input: the real and the generated MVTS samples. To simplify the representation for our discussion about the discriminator, we denote ${X_{n}}$ as a uniform set of inputs. Through feeding $C_{n}$ into \textit{D}, the discriminator not only produces judgments about whether the data is synthetic or real but also evaluates the correspondence of the synthetic sample to its conditional information. Finally, the binary cross-entropy loss calculated between the prediction and the ground truth is used to update the parameters of both the generator and the discriminator with the back-propagation algorithm.
    
    So far, we have comprehended the structures and functionalities of the generator and discriminator. Next, we will define the objective function used for optimizing the discriminator and the generator. In our framework, the objective function is divided into two parts, including the generator loss and the discriminator loss. First, the discriminator loss which is calculated as the cross-entropy between the ground-truth and outputs of a discriminator, is defined as: 
    
    \begin{equation}
    \begin{footnotesize}
    \label{eq:D_loss}
    Loss_{D}(X_{n}|C_{n}, y_{n}) = -\text{CE}\Big(D(X_{n}|C_{n}), y_{n}\Big)
    \end{footnotesize}
    \end{equation}
    
    \par
    In this equation, $X_{n}$ is the set of inputs of the discriminator, and $C_{n}$ is the conditional vector. $D(X_{n}|C_{n})$ returns the probability of $X_{n}$ being a real or synthetic sample by taking $X_{n}$ and $C_{n}$ as inputs. Note that ${X_{n}}$ in this equation is composed of two different types of data sources:
    
    \begin{equation} \label{eq:X_n}
    \begin{footnotesize}
    \begin{aligned}
    \textit{$X_{n}$}\!=\!\!
    \begin{cases}
    \textit{$X_{n}$} & \textbf{if~}inputs\ are\ real\ samples,
    \\
    \mbox{G}(Z_{n}|C_{n}) & \textbf{if~}inputs\ are\ synthetic\ samples.
    \end{cases}
    \end{aligned}
    \end{footnotesize}
    \end{equation}
    
    \par
    Correspondingly, the $y_{n}$ takes two different values, dependent upon the source of the sample in $X_{n}$,
    
    \begin{equation} \label{eq:y_n}
    \begin{footnotesize}
    \begin{aligned}
    \textit{$y_{n}$}\!=\!\!
    \begin{cases}
    \textit{\textbf{1}} & \textbf{if~}inputs\ are\ real\ samples,
    \\
    \textit{\textbf{0}} & \textbf{if~}inputs\ are\ synthetic\ samples.
    \end{cases}
    \end{aligned}
    \end{footnotesize}
    \end{equation}
    
    The generator loss is also formulated with cross entropy as below:
    
    \begin{equation} \label{eq:G_loss}
    \begin{footnotesize}
    Loss_{G}(Z_{n}|C_{n}) = -\text{CE}\bigg(D\Big(G(Z_{n}|C_{n})|C_{n}\Big), \textbf{1}\bigg)\\
    \end{footnotesize}
    \end{equation}
    
    \noindent where the input $G(Z_{n}|C_{n})$ is the synthetic samples, and its corresponding predictions are $D\Big(G(Z_{n}|C_{n})|C_{n}\Big)$. In calculating the loss of the generator, the label of a synthetic sample is held constant as $\textbf{1}s$, since the goal of the generator is to generate realistic-enough samples such that the discriminator can no longer distinguish them from the real samples.
    

\section{Experiments}
    Despite the model that can be optimized according to the objective function defined in Section~\ref{sect:methodology}, the loss cannot objectively reflect the convergence of the training progress or the quality of generated samples \cite{arjovsky2017wasserstein}. To determine when to stop training models is a well-known and up in the air question in the GAN study. Unlike many image-based GAN projects, such as deepfake \cite{chan2019everybody} and GauGAN \cite{park2019gaugan}, the visual verification of synthetic time series as outputs does not give us much evidence as to whether the generated data are realistic or not. In this section, we present three types of evaluation methods, in both qualitative and quantitative metrics, to verify the effectiveness and correctness of the CGAN model. 

    \subsection{Experimental settings}\label{sect:experiment:settings}

    We have implemented our model using the TensorFlow 2.0 library \cite{tensorflow2015-whitepaper}, and the code can be found at our repository for the project\footnote{https://bitbucket.org/gsudmlab/mvts-gann}. After we explored various settings based on the defined objective function, we found that using the Adam Optimizer for the generator and the Gradient Descent Optimizer for the discriminator produced our best results. Moreover, we concluded our parameter settings of the learning rate to $0.1$, the LSTM hidden size to $100$, and the batch size to $32$. The model is trained with $300$ epochs, and intermediate models are saved every five epochs. In Sections~\ref{sect:experiment:statistics} and \ref{sect:experiment:aa}, we utilize $1,254$ real flare samples in partition $1$ of SWAN-SF and $1,254$ synthetic samples generated by the CGAN model in our evaluations. In Section~\ref{sect:experiment:svm} though, we utilize the entire partition $1$, including $1,254$ real flares and $72,238$ no-flare samples, and generate $70,984$ synthetic flare samples to balance the training dataset. More details regarding the experimental setup will be presented in that section.
    
    \subsection{Evaluation using the distributions of statistical features}\label{sect:experiment:statistics}
    
    To provide a statistical evaluation for our model, we utilize the statistical features extracted from the time series data, including the mean, median, and standard deviation statistics. For visualization, we construct frequency distributions of these values for both the real input data and the synthetic data produced by the trained generator through binning the values into $20$ equal-width bins. It is our assumption that if the distributions of the statistic values for both the real data and generated data are similar, then the generated samples should be similar enough to perform well for our final task of producing minority class samples for model training.
    
    \begin{figure} [hbt!]
        \centering
        \includegraphics[scale=0.6]{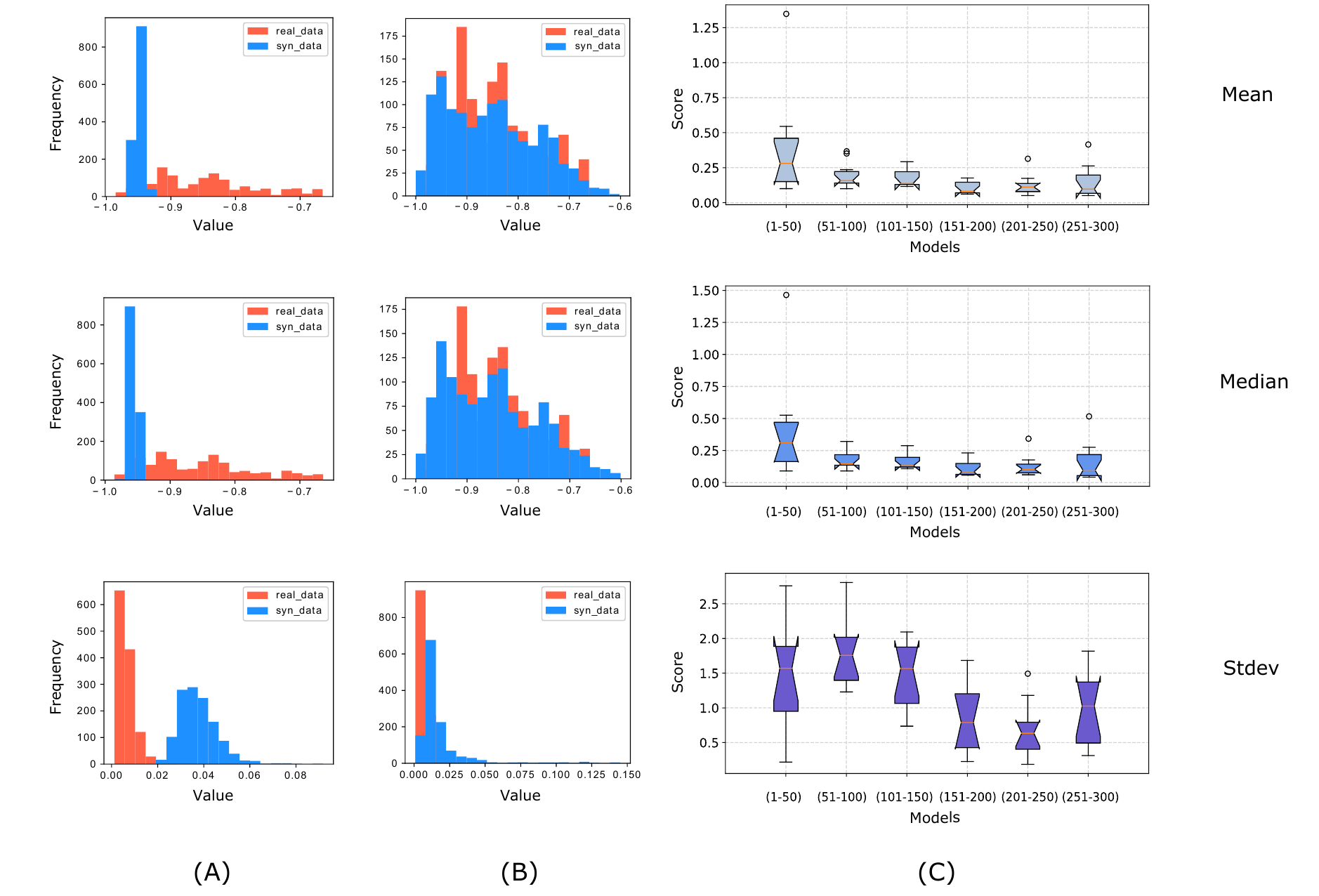}
        \caption{Columns (A) and (B) show the distributions of three features, including mean, median, and standard deviation, of TOTUSJH at two intermediate epochs with a bin size of $20$. Red bars stand for the real data, and blue bars stand for the synthetic data. Specifically, (A) is the result of the model at the $50$th epoch, and (B) is the result of the model at the $250$th epoch. Column (C) shows the distributions of KL divergence scores calculated by comparing distributions of synthetic samples and real samples across all intermediate models divided into six groups.}
        \label{fig:KL_Evaluation}
    \end{figure}
    
    We have conducted statistical feature evaluation on all the physical parameters, but for brevity, we present only the results of TOTUSJH. The columns (A) and (B) in Fig.~\ref{fig:KL_Evaluation} show the distributions of three features evaluated based on two intermediate models saved in the training process. (A) is resulted after training 50 epochs, which is the last model of the first group (1-50 epoch), and (B) is resulted after training 250 epochs, which is the last model of the fifth group (201-250 epoch). From (A) to (B), it can be found that the generator can gradually produce time series which have similar statistic attributes to the real data. Additionally, we calculate the Kullback–Leibler (KL) divergence between distributions of real data and generated samples with all features at different epochs. We observe, as shown in column (C) of Fig.~\ref{fig:KL_Evaluation}, that the KL divergences of all three features are decreasing as training progresses, which means that the two distributions are getting more and more similar as the model evolves. We found that models between the 201-250 epochs achieve the best performance, with lower KL divergence for the mean, median, and standard deviation value distributions. We also see that the variance between the results produced by intermediary models trends downward until we surpass the 250 epoch mark. This is regarded as the first criterion of the model selection in this study.

    \subsection{Evaluation using Adversarial Accuracy}\label{sect:experiment:aa}

    The Adversarial Accuracy, as formulated in Eq.~\ref{eq:Adversarial_Accuracy}, is put forward by Yale \cite{yale2019assessing}, which is used for comparing the similarity of two sets of data samples.
    
    \begin{equation} \label{eq:Adversarial_Accuracy}
    \begin{footnotesize}
    AA_{TS} = \frac{1}{2}(\frac{1}{n}\sum_{i=1}^{n}\boldsymbol{1}(d_{TS}(i)>d_{TT}(i)) + \frac{1}{n}\sum_{i=1}^{n}\boldsymbol{1}(d_{ST}(i)>d_{SS}(i)))
    \end{footnotesize}
    \end{equation}
    
    In the definition, the variable $T$ stands for the set of real data, and the variable $S$ stands for the set of synthetic data. For calculating $\textbf{1}(d_{TS}(i) > d_{TT}(i))$, each real sample from $T$ are compared with all synthetic data points in $S$ to calculate the shortest distance $d_{TS}(i)$, and compared with all the other real data points in $T$ to calculate the shortest distance $d_{TT}(i)$. The most straightforward distance metric is the Euclidean distance which is our choice in this study as well. The shortest distance generally means the highest similarity between two data points. If $d_{TS}(i) > d_{TT}(i)$, it means no synthetic data point can be found that is more similar to the current real data point than other real data points. Otherwise, a synthetic data point, which is more similar to the current real data point, can be found. The $d_{TS}(i) < d_{TT}(i)$ indicates a realistic samples is generated. The second part, $\textbf{1}(d_{ST}(i) > d_{SS}(i))$, is implemented in a similar manner, except that here each synthetic sample will be compared with not only all the real samples but also all the other synthetic samples. Overall, the best balance result of Adversarial Accuracy should equal 0.5, which implies the generator can generate realistic samples.
    
    \begin{figure} [hbt!]
        \centering
        \includegraphics[scale=0.55]{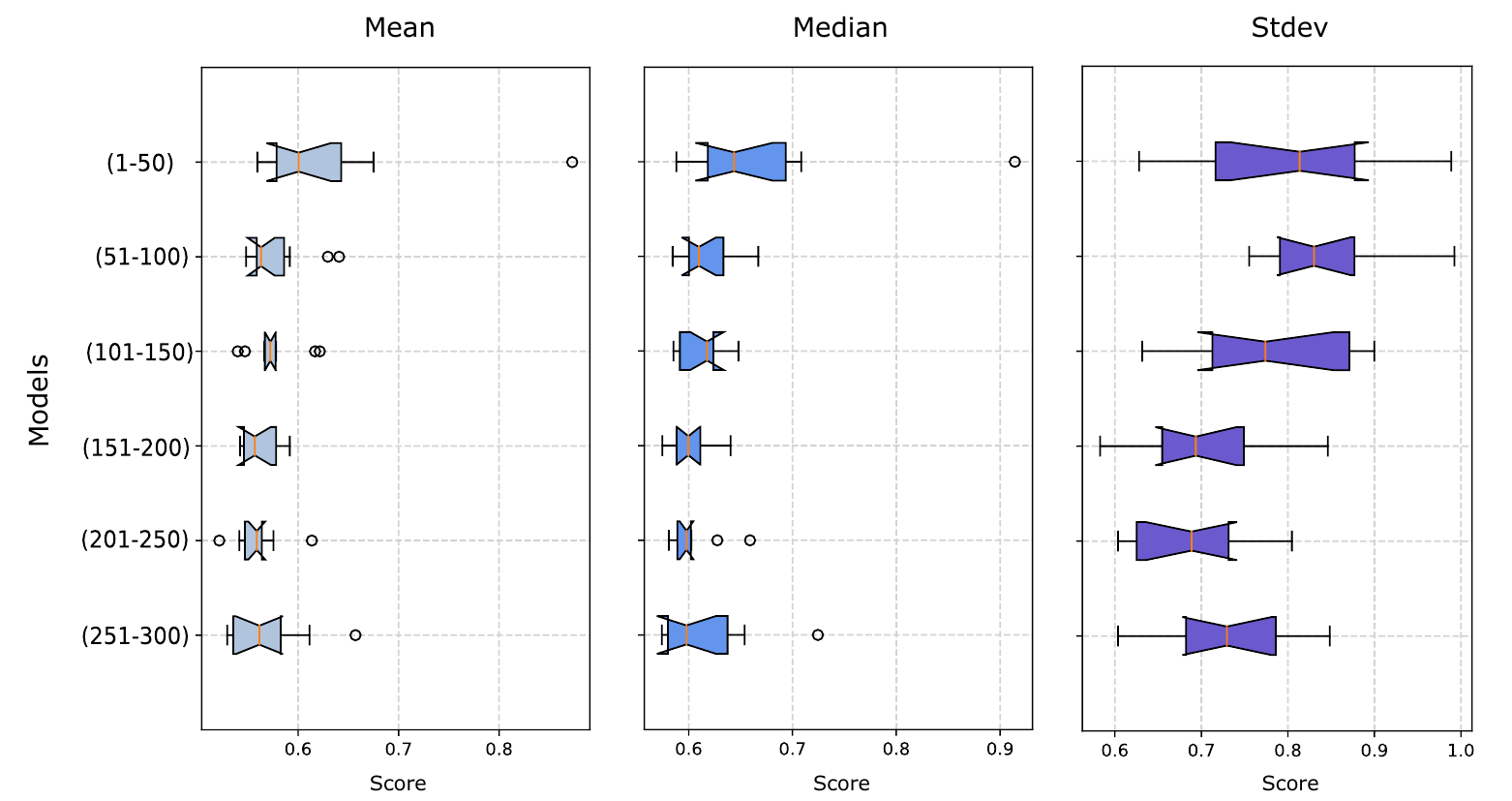}
        \caption{The box plots show the distributions of Adversarial Accuracy of three features, including mean, median, and standard deviation, of TOTUSJH evaluated with all intermediate models by divided into six groups. This metric is regraded as the second criterion for exploring the model selection.}
        \label{fig:AA_Evaluation}
    \end{figure}
    
    In our experiments, each time series is represented by extracting statistical attributes of the mean, median, and standard deviation. Then, we utilize each attribute to calculate the Adversarial Accuracy among all samples using Eq.~\ref{eq:Adversarial_Accuracy} with all intermediate models. Fig.~\ref{fig:AA_Evaluation} shows the results of the three statistical features of TOTUSJH. Through observing the box plots, we find that the models between 201 to 250 epochs can achieve $0.55$ in mean, $0.60$ in median, and $0.68$ in standard deviation on average, which shows that the CGAN model can generate synthetic samples by maintaining a good balance between underfitting and overfitting with real samples. Moreover, the Adversarial Accuracy results have a consistent conclusion with the KL divergence experiment in Section~\ref{sect:experiment:statistics} regarding the model selection.
    
    \subsection{Evaluation using SVM classifiers}\label{sect:experiment:svm}
    
    In this section, we evaluate how well the generated data remedy the class-imbalance issue in classification of flaring and non-flaring instances of SWAN-SF. First, we train the generator on partition $1$ of SWAN-SF, with the four magnetic field parameters mentioned in Section~\ref{sect:related-work:swan-sf}. The outputs of the generator are the synthetic multivariate time series most similar to the actual multivariate time series of flares. Then, we move to train two classifiers: one on the highly imbalanced real data without any change, and the other, on the data that is made balanced by adding synthetic samples. Of course, the synthetic samples are only used for the purpose of training, and the validation and test sets are made entirely of real data. We consider the former as the baseline. The only difference between the two classification strategies is that they are trained with different training datasets.
    
    Both classifiers are trained on partition 1 and evaluated on partitions 2, 3, and 5. Partition 4 is used for tuning hyperparameters. We choose the SVM as the standard classifier for both classification tasks with the same settings that conclude the hyperparameters C and $\gamma$ to 0.25 and 0.25. For preprocessing of SWAN-SF, we linearly transform all five partitions to the range $[-1, 1]$ for training the CGAN model and SVM classifiers, and impute Missing data with the linear interpolation.
    
    Considering the flare forecasting problem that we approach is in fact a rare-event classification task, choosing a proper evaluation is important. From years of exploration, domain experts have come to agree on the effectiveness of two metrics, namely the \textit{true skill statistic} ($TSS$) \cite{hanssen1965relationship} and the updated \textit{Heidke skill score} ($HSS2$) \cite{balch2008updated} as shown in Eq.~\ref{eq:tss} and Eq.~\ref{eq:hss2}.
        
    \begin{equation}\label{eq:tss}
    \begin{footnotesize}
        TSS = \frac{tp}{tp + fn} - \frac{fp}{fp + tn}
    \end{footnotesize}    
    \end{equation}
    
    \begin{equation}\label{eq:hss2}
    \begin{footnotesize}
        HSS2 = \frac{2 \cdot ((tp \cdot tn) - (fn \cdot fp))}{(tp + fn) \cdot (fn + tn) + (fp + tn) \cdot (tp + fp)}
    \end{footnotesize}
    \end{equation}

    \begin{figure} [hbt!]
      \centering
    \includegraphics[scale=0.45]{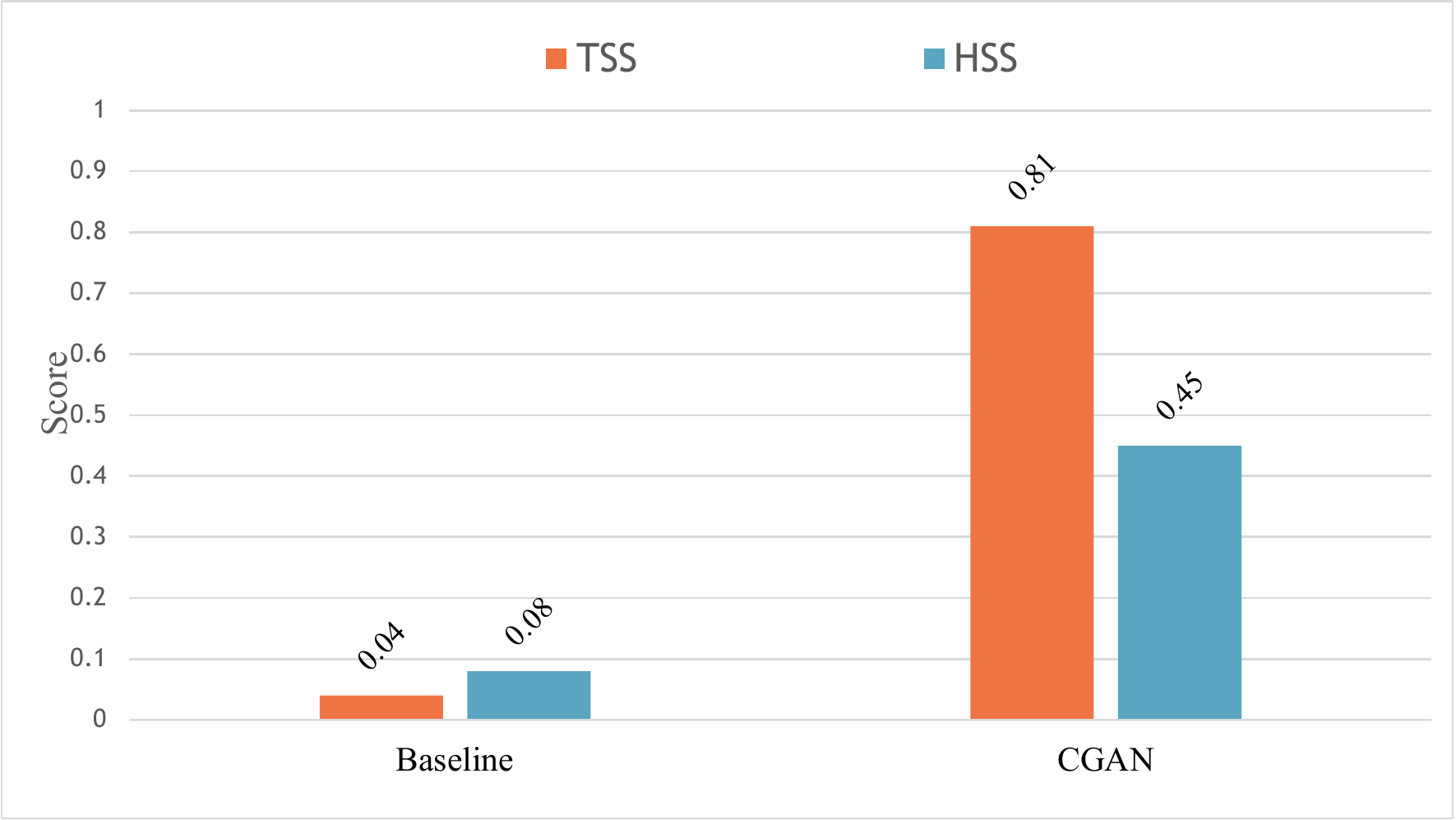}
      \caption{This is the evaluation of flare forecasting based on SVM classifiers. The Baseline experiment (Left) is trained on partition 1 (imbalanced) of SWAN-SF, and the CGAN experiment (Right) is trained on the balanced partition 1 by adding synthetic samples generated by the CGAN model. Both experiments are evaluated on the same test dataset: partitions 2, 3, and 5 of SWAN-SF. From the results, we find notable improvements in terms of TSS and HSS. This experiment shows the CGAN can be considered as an effective remedy for tackling the class imbalance issue.}
      \label{fig:SVM_Evaluation}
    \end{figure}
    
    As the forecasting results reported in Fig.~\ref{fig:SVM_Evaluation}, we observe that the performance of the classifier trained on the training dataset which was made balanced using our generated synthetic samples is remarkably higher than that of the baseline classifier, by both metrics , TSS and HSS2. This observation confirms that the model generally performs best when classes in the training dataset are roughly equal in size. Specifically, the CGAN classifier results in an over 20-fold improvement compared to the baseline experiment in terms of TSS (an increase from 0.04 in average to 0.81). Moreover, the HSS2 improves an over 5-fold improvement from 0.08 to 0.45. The CGAN experiment results show that the CGAN model can successfully capture statistical features of real MVTS samples by learning the data distribution and, therefore, generating realistic MVTS samples. Overall, the CGAN model can be regarded as an effective remedy for tackling the class imbalance issue.


\section{Conclusion \& Future Work}
    
    In this project, we utilized the conditional generative adversarial network (CGAN) to overcome the class imbalance issue with a multivariate time series dataset. We generated synthetic samples and evaluated them by conducting two sets of experiments: First, experiments based on the data distributions of statistical features and the Adversarial Accuracy verified that synthetic samples are indeed similar to the real data. Next, our classification experiment showed that generating synthetic samples of the minority class in order to balance the training dataset can remarkably boost the performance of classification models. Therefore, we concluded that the CGAN method can be considered as an effective remedy for tackling the class imbalance issue in flare forecasting.
    
    Of course, this study is only the first attempt towards generating a reliable synthetic dataset with meaningful physical features. There are still many aspects to be improved upon, such as incorporating an advanced loss function of Wasserstein GAN, introducing the representation learning of Info GAN, or exploring more complex structures of the generator and the discriminator.As our future work, we plan to compare the presented approach with other class-imbalance remedies including simple oversampling and undersampling strategies, in order to provide more insights into the effectiveness of the CGAN approach. Furthermore, exploring and interpreting the meaning of synthetic samples from the astrophysics point of view is also a worthwhile topic that we wish to investigate in the future, by collaborating with domain experts.


\section*{Acknowledgment}
    
    It will be provided at publication.
    
\balance
\bibliographystyle{IEEEtran}
\bibliography{bibliography} 

\end{document}